\documentclass[letterpaper, 10 pt, conference]{ieeeconf}  

\IEEEoverridecommandlockouts                              

\usepackage{graphics} 
\usepackage{epsfig} 
\usepackage{mathptmx} 
\usepackage{times} 
\usepackage{amsmath} 
\usepackage{amssymb}  

\usepackage{amsfonts}
\usepackage{booktabs}       
\usepackage{xcolor}
\usepackage{soul}

\title{\LARGE \bf
Row-wise LiDAR Lane Detection Network\\with Lane Correlation Refinement
}

\author{Dong-Hee Paek$^{1}$, Kevin Tirta Wijaya$^{2}$, and Seung-Hyun Kong$^{*1}$
\thanks{$^{*}$corresponding author}
\thanks{$^{1}$Dong-Hee Paek and Seung-Hyun Kong are with the CCS Graduate School of Mobility, Korea Advanced Institute of Science and Technology, 193, Munji-ro, Yuseong-gu, Daejeon, Republic of Korea
        {\tt\small \{donghee.paek, skong\}@kaist.ac.kr}}%
\thanks{$^{2}$Kevin Tirta Wijaya is with the Robotics Program, Korea Advanced Institute of Science and Technology, 291, Daehak-ro, Yuseong-gu, Daejeon, Republic of Korea
        {\tt\small kevin.tirta@kaist.ac.kr}}
}

\begin{document}

\maketitle
\thispagestyle{empty}
\pagestyle{empty}

\begin{abstract}
Lane detection is one of the most important functions for autonomous driving.
In recent years, deep learning-based lane detection networks with RGB camera images have shown promising performance.
However, camera-based methods are inherently vulnerable to adverse lighting conditions such as poor or dazzling lighting.
Unlike camera, LiDAR sensor is robust to the lighting conditions.
In this work, we propose a novel two-stage LiDAR lane detection network with row-wise detection approach.
The first-stage network produces lane proposals through a global feature correlator backbone and a row-wise detection head.
Meanwhile, the second-stage network refines the feature map of the first-stage network via attention-based mechanism between the local features around the lane proposals, and outputs a set of new lane proposals.
Experimental results on the K-Lane dataset show that the proposed network advances the state-of-the-art in terms of F1-score with 30\% less GFLOPs.
In addition, the second-stage network is found to be especially robust to lane occlusions, thus, demonstrating the robustness of the proposed network for driving in crowded environments.
\end{abstract}

\section{INTRODUCTION}

To be able to navigate from a point to another, an autonomous driving agent needs to plan a safe and efficient route according to the environmental conditions.
Therefore, the ability of perceiving the environment through raw sensor measurements is crucial for the autonomous driving.
One of the perception tasks for autonomous driving is the lane detection task, where the autonomous driving agent needs to detect the location of lane lines on the roads.

Extensive studies have been conducted on the lane detection task, particularly with the RGB camera sensors.
In the earlier days, rule-based and heuristic systems have been developed to provide lane detection capability in limited predefined environments \cite{heu_cam_1}\cite{heu_cam_2}\cite{heu_cam_3}.
Recently, data-driven approaches become popular owing to the advancements of deep learning.
Various neural networks for camera-based lane detection have been developed \cite{ufast}\cite{laneaf}\cite{condlane}, with promising accuracy in most of driving conditions.

However, RGB camera has an inherent weakness towards harsh lighting conditions such as low or dazzling light.
This is evident in the widely-used CULane benchmark \cite{scnn}, where the performance degradation of various camera-based lane detection networks occurs in the dazzling light and dark.
As an autonomous driving agent needs to be robust in various driving conditions, the vulnerability of the existing camera-based lane detection methods towards adverse lighting conditions need to be resolved.

One possible solution to the problem is to use Light Detection and Ranging (LiDAR) sensor.
Since the LiDAR sensor emits infrared signals, which is hardly interfered by the visible light, adverse lighting conditions do not affect its measurement capability significantly.
Moreover, unlike camera images, the LiDAR point cloud does not require a bird's eye view (BEV) projection for motion planning, which often causes lane line distortions.

Despite the several advantage of the LiDAR sensor, only a handful of studies have proposed deep learning-based LiDAR lane detection methods.
This is largely due to the absence of publicly-available LiDAR lane detection datasets.
As seen in the deep camera-based lane detection field, the majority of the lane detection networks are developed after the publication of open lane detection datasets such as \cite{scnn}\cite{tusimple}.

Recently, \cite{klane} opens a large-scale LiDAR lane detection dataset, K-Lane, to the public, along with a segmentation-based LiDAR lane detection network (LLDN) baseline.
The baseline consists of a projection network, a global feature correlator (GFC), and a segmentation head, which generates a feature map in the BEV format, extracting features via global correlation, and predicting lanes per grid, respectively.
While a segmentation-based lane detection network is capable of producing lane detection of various shapes, it is computationally expensive due to the need of processing each grid of the final feature map through shared multi-layer perceptron (MLP).

In this work, we propose a novel LiDAR lane detection network that is computationally efficient and especially effective to the severe occlusion cases.
We set the lane detection problem as a row-wise prediction task instead of a semantic segmentation task, so that the detection head of the network performs row-wise MLP operations instead of grid-wise MLP operations.
The row-wise formulation leads to a significantly less computational cost than the prior work \cite{klane}, with about 30\% less GFLOPs. 

Furthermore, we design an additional second-stage network that refines the output feature map of the first-stage network via correlation of features around the lane proposals.
The correlation process is implemented with the dot-product attention mechanism, which allows the network to exchange information between the features of the lane proposals globally.
As the lane lines often have some degree of regularity in terms of shapes and distances between each other, such global correlation process should be advantageous, for example, in detecting lane lines that are occluded by neighboring vehicles.
The row-wise two-stage approach enables our proposed network to achieve the state-of-the-art (SOTA) performance with less computational cost compared to the existing baselines.

In a summary, our contributions are as follows:
\begin{itemize}
    \item {We propose a new technique in LiDAR lane detection through row-wise lane prediction. This technique is more computationally efficient compared to existing segmentation-based LiDAR lane detection techniques.}
    \item {We design a two-stage LiDAR lane detection network that first predicts lane proposals, and then refines the first-stage feature map via attention-based mechanism between the lane proposals. The refined feature map is then used to predict the final lane detection proposals.} 
    \item {We demonstrate the excellent performance of the proposed network; the proposed network achieves SOTA performance in K-Lane with an overall F1-score of 82.74\% reducing the GFLOPs by about 30\%. While the proposed network achieves a slight overall F1-score improvement (i.e., 0.62\%) over the prior SOTA network, the proposed network largely improves the F1-score (i.e., 3.24\%) for the severe occlusion cases. Thus, the proposed network enables the robust lane detection required for safe autonomous driving in congested traffics where performance degradation has previously been significant.}
\end{itemize}

The rest of this paper is organized as follows: Section II introduces existing works related to our study, Section III details our proposed LiDAR lane detection network, Section IV discusses the experimental results on the K-Lane dataset, and Section V draws conclusion this study.

\section{Related Works}
In this section, we discuss existing works that are related to our study.
We start with a general review of the more matured camera-based lane detection.
Then we discuss further on the row-wise lane detection methods.
Finally, we cover the existing LiDAR lane detection methods.

\subsection{Camera-based Lane Detection}
Traditional camera-based lane detection methods heavily rely on rule-based systems that require various predefined variables such as intensity thresholding \cite{heu_cam_1}\cite{heu_cam_2}\cite{heu_cam_3}.
As the deep learning field becomes more matured, various camera lane detection networks emerge, with promising accuracy in various driving conditions.
Most deep learning-based camera lane detection networks utilize convolutional neural network (CNN) as their backbone feature extractor, and a task-specific detection head.

In \cite{laneatt}, the detection head is designed to perform anchor-based lane predictions and coordinate offset predictions. 
In \cite{ufast}, the detection head produce two affinity-field maps that represent the locations of lane lines.
In \cite{condlane}, the detection head is conditioned to predict row-wise lane proposals before being further processed by a post-processing algorithm.
Compared to other approaches, row-wise lane detection often perform faster while maintaining good accuracy.
As such, we design our LiDAR lane detection network with a row-wise paradigm.

\subsection{Row-wise Lane Detection}
Row-wise lane detection is proposed by \cite{fastdraw}, in which lane lines are detected through predicting the lane location probability of each row.
That is, for each row, the location of the lane line is determined as the column of which the lane probability is the highest.
Unlike segmentation-based lane detection, row-wise lane detection is based on geometric prior which picks only one location per each lane, so that this method may robust to false alarm.
Furthermore, \cite{ufast} modify the argmax operation into taking the sums of each index of the column weighted by its lane probability to enable the gradients to flow through the lane structure loss.

\subsection{LiDAR Lane Detection}
In the early LiDAR lane detection methods, lane lines are  detected through an intensity thresholding operation with additional heuristics.
These methods often incorporate additional algorithms such as Kalman Filter \cite{heu_lidar_1}, polar coordinates operation \cite{heu_lidar_2}, or clustering with DBSCAN \cite{heu_lidar_3}.
However, heuristic methods are not adaptive towards diverse driving conditions due to the need for predefining numerous thershold variables.

More recently, several studies are starting to incorporate deep learning to their LiDAR lane detection methods.
In \cite{lane_fusion}, the front view camera images are combined with the 2D BEV images from the LiDAR point cloud to improve the lane detection performance.
In \cite{deeplidar_egolane}, a CNN backbone is utilized to detect the ego lane lines in the highways.
In \cite{klane}, a segmentation-based neural network with global feature correlator backbone is used to perform LiDAR lane detection under various environments in the K-Lane dataset.

\begin{figure*}[htb!]
\centering
\includegraphics[width=1.98\columnwidth]{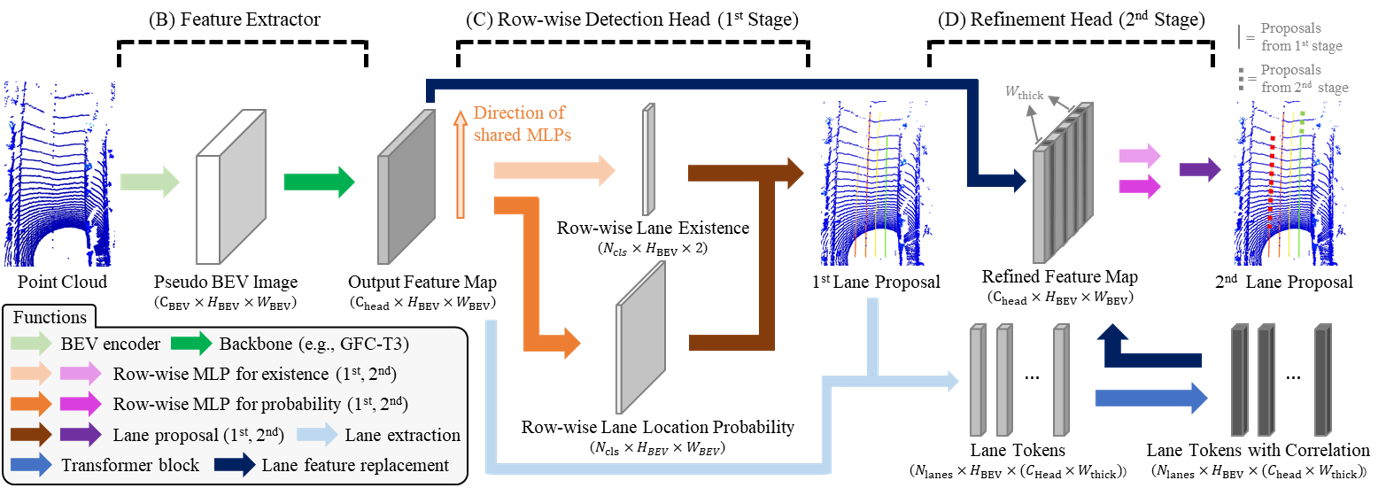}
\caption{Overall structure of the proposed network.}
\label{fig1}
\end{figure*}

\section{Method}
In this section, we describe the proposed network in details.
Firstly, we provide an overview of the structure of the proposed network.
Then, we explain in details about the components of the network: the feature extractor, the row-wise detection head, and the refinement head.
Finally, we express the loss function that is used to train the network.

\subsection{Two-stage Row-wise Lane Detection Network}
As shown in Fig. \ref{fig1}, we design the proposed lane detection network in two-stage detection network with a row-wise approach.
First, a feature extractor backbone takes the raw point cloud data as an input.
The feature extractor then encodes the raw point cloud into a pseudo bird-eye-view (BEV) image, which is further processed by a global feature correlator to produce the output feature map.
This output feature map is utilized by the first stage row-wise detection head to predict two values: row-wise existence and row-wise probability.
Through the row-wise existence and location probability predictions, we can obtain the first-stage lane detection proposals.

To further improve the lane detection accuracy, we use a second-stage refinement head.
The refinement head collects features around the location of the lane proposals, which we term as lane tokens.
The lane tokens are then processed through the attention-based mechanism, resulting in new lane tokens with correlation information.
The new lane token features are used to replace the original lane features in the output feature map of the first stage network, resulting in a refined feature map.
Finally, the refinement head produce new and more accurate lane detection proposals through another row-wise existence and probability predictions from the refined feature map.

\subsection{Feature Extractor}
The feature extractor is responsible for encoding the point cloud raw data into an output feature map that is used by the detection head to make the final predictions.
It is composed of two parts: the BEV encoder and the global feature correlator (GFC) backbone.
Given a point cloud $\textbf{P} = \{ \textbf{p}_1, \textbf{p}_2, ..., \textbf{p}_n \}$, where $\textbf{p}_i \in \mathbb{R}^{(3+C)}$ is a point in the 3D space with $C$ additional features such as intensity and reflectivity, the feature extractor first encode the raw point cloud data into a pseudo BEV image of size $C_{BEV} \times H_{BEV} \times W_{BEV}$, where $C_{BEV}$ is the number of feature channels, $H_{BEV}$ is the number of rows, and $W_{BEV}$ is the number of columns.

After obtaining the pseudo BEV image, the backbone then learn important features through global feature correlator.
This results in the final output feature map of size $C_{head} \times H_{BEV} \times W_{BEV}$.
We utilize the same feature extractor as seen in previous state-of-the-art network \cite{klane} as our focus in this work is on the detection head with the two-stage row-wise formulation.

\subsection{Row-wise Detection Head}
\begin{figure}[b!]
\centering
\includegraphics[width=1.0\columnwidth]{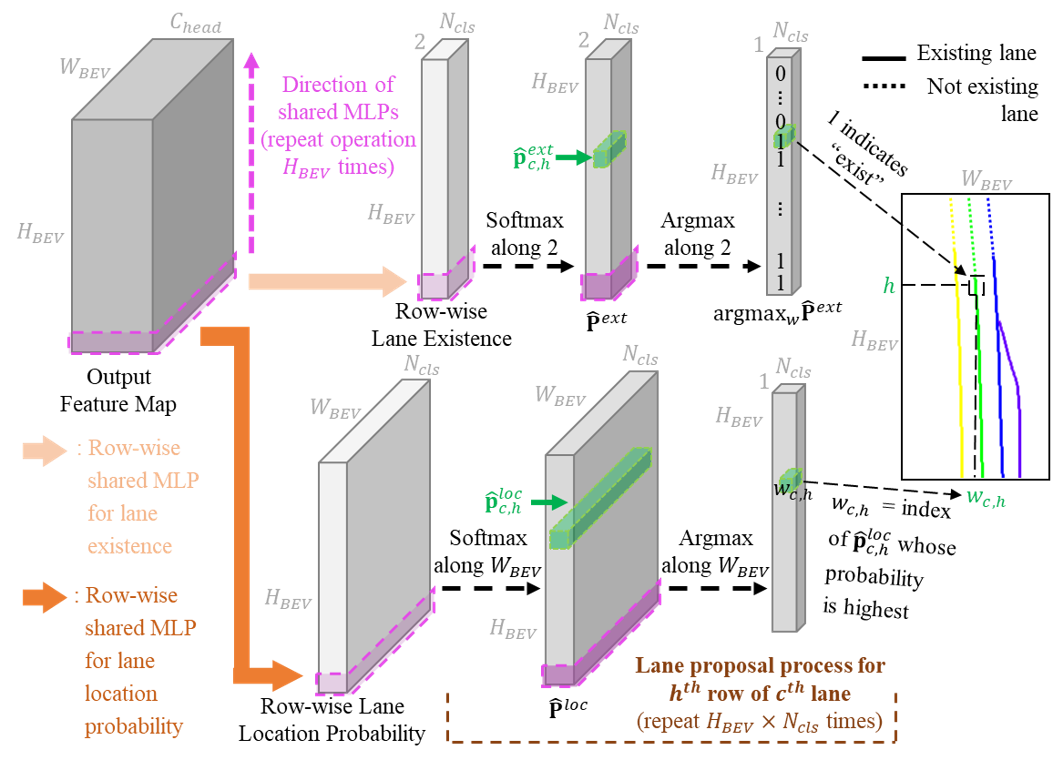}
\caption{Detailed process of row-wise detection head.}
\label{fig2}
\end{figure}

The row-wise detection head uses the final output feature map as an input and produce two predictions: the row-wise lane existence and the row-wise lane location probability.
To do so, we leverage the fact that lane lines from a LiDAR scan have almost no shape distortion along the BEV map rows, thus, it is suitable to utilize shared-MLPs.
As shown in Fig. \ref{fig2}, the MLPs are shared along the rows of the feature map, that is, each row in the feature map is considered as an individual feature vector (colorized as purple in Fig. \ref{fig2}) to be processed by the same MLPs.

We use two distinct shared-MLPs, one for predicting whether a lane line class exists on the road (row-wise lane existence), the other for predicting the horizontal (column-wise) location on which the lane line resides on (row-wise location probability) as shown in Fig. \ref{fig2}.
Specifically, the row-wise existence MLPs output is an $N_{cls} \times H_{BEV} \times 2$ logits feature map, where $N_{cls}$ is the number of lane line classes, and 2 is the number of parameters for predicting  the probability of existence (i.e., exist or not), with column 0 represent not-exist flag, while column 1 represent exist flag.
The row-wise probability MLPs output is an $N_{cls} \times H_{BEV} \times W_{BEV}$ logits feature map.
To obtain the row-wise existence and location probability, we apply softmax function along the column to the logits of feature maps.
The existence and the location of lane line of class $c$ on row $h$ is determined through the argmax function along the rows.
Note that while each rows contains lane location probability, only rows with a positive existence prediction are considered as the first-stage lane proposals.

\subsection{Refinement with Lane Correlation}
After obtaining the first lane proposals from the detection head, the network further improve its predictions through the refinement head as shown in Fig. \ref{fig1} (D).
The refinement head first collects features from the final output feature map based on the location of lane proposals, resulting in a set of lane tokens.
However, a lane class that does not have enough information may adversely affect the correlation operation, so we only extract features of lane classes that have more than a certain amount of positive existence along the entire rows.

To be specific, suppose that the lane existence predictions are $\hat{\textbf{P}}^{ext} \in \mathbb{R}^{N_{cls} \times H_{BEV} \times 2}$ as described in the previous subsection.
We collect features of lane class $c$ iff., 
\begin{equation}
    \frac{1}{H_{BEV}}\sum_{h}^{H_{BEV}} argmax_{w}(\hat{\textbf{{p}}}^{ext}_{c,h}) >T_{ext},
    \label{eq:req}
\end{equation}
i.e., we only consider the lane classes that exists on more than $T_{ext}\cdot H_{BEV}$ number of rows.

The lane tokens have a size of $N_{lanes} \times H_{BEV} \times (C_{head} \times W_{thick})$, where $N_{lanes}$ is the number of extracted lanes following the requirement stated in equation \ref{eq:req}, and $W_{thick}$ is the {thickness of the lane} that we {extract} when collecting the tokens.
For an example, {the thickness of three} will result in a token that is constructed from the feature vector at the same coordinate as the lane proposal, and two other feature vectors from the neighboring columns.

The collected lane tokens are then processed by a transformer encoder block \cite{transformer} to produce refined feature vectors.
Intuitively, the transformer encoder block tries to find important features in the lane tokens while considering the interaction between feature vectors of different lanes via correlation (i.e., dot product of query and key).
Such mechanism may be advantageous considering that most lane lines are constructed with a certain degree of regularity in terms of shape and distance between each other.

The lane tokens are then returned back to their original coordinates on the output feature map to create a refined feature map.
This refined feature map is then used to predict the final row-wise existence and row-wise probability through two distinct shared-MLPs as in the first stage prediction.
As such, the lanes, which are difficult to be detected in the first proposal, can be detected through interaction with other lane lines.
This hypothesis is backed by both quantitative and qualitative enhancement, especially for hard cases such as severe occlusions. 

\begin{figure*}[tb!]
\centering
\includegraphics[width=1.95\columnwidth]{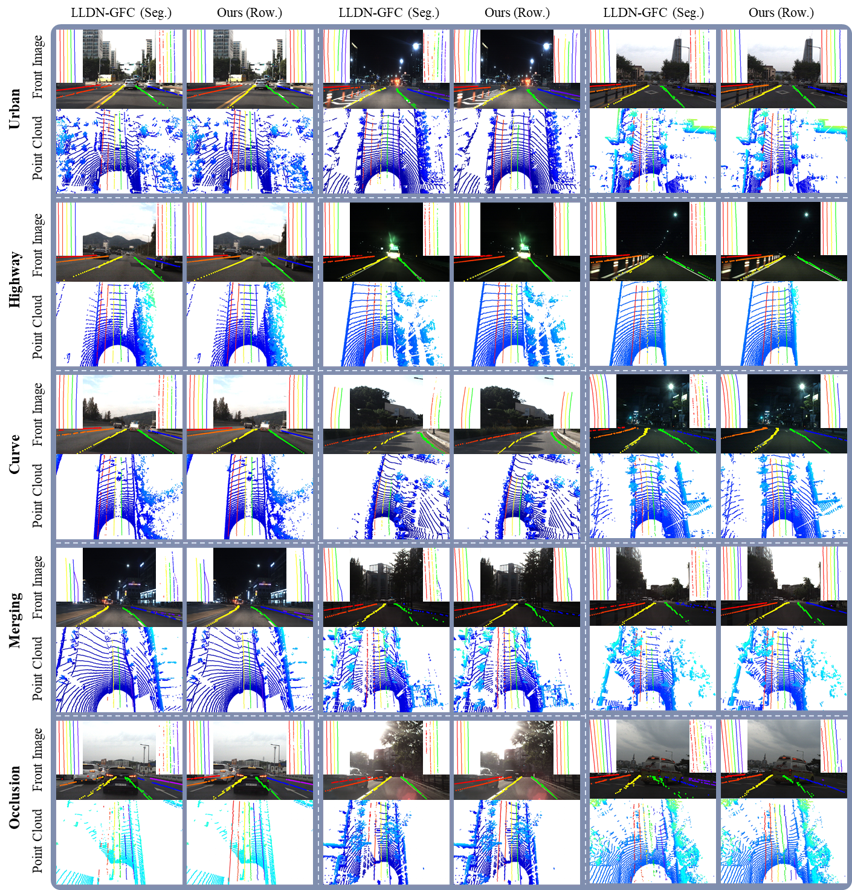}
\caption{Qualitative comparison of segmentation-based (SOTA prior work) and row-wise-based (ours): We show lane detection results in a total of five different conditions, i.e. urban, highway, curve, merging, and occlusion. We project the lane proposals onto the RGB camera image, where the BEV label is shown on the top-left corner, and the BEV prediction is shown on the top-right corner. We also visualize the lane proposals on the point cloud under its corresponding RGB camera image.}
\label{fig3}
\end{figure*}

\begin{table*}[ht]
    \centering
        \caption{{Comparison of F1-score on Various Road Conditions} \& GFLOPs}
    \label{tab:exp_perform}
    \begin{tabular}{cccccccccccccc}
    \toprule
         Model & Overall & Daylight & Night & Urban & Highway & Curve & Merging & \multicolumn{5}{c}{Num. of Occluded Lines} & GFLOPs \\
        \cmidrule(lr){9-13}
         & & & & & & & & 0 & 1 & 2 & 3 & 4-6 \\
         \cmidrule(lr){1-14}
         LLDN-GFC \cite{klane}   & 82.12 & 82.22 & 82.00 & \textbf{81.75} & 82.55 & \textbf{78.05} & \textbf{81.08} & 82.97 & 81.43 & 81.28 & 78.67 & 75.92 & 558.0 \\
         Ours (1-stage)     & 82.37 & 82.04 & 82.75 & 81.14 & 83.85 & 76.49 & 79.78 & 83.33 & 81.52 & 81.57 & 77.91 & 76.36 & \textbf{385.1} \\
         Ours (2-stage)     & \textbf{82.74} & \textbf{82.58} & \textbf{82.92} & 81.64 & \textbf{84.05} & 76.16 & 79.92 & \textbf{83.44} & \textbf{81.87} & \textbf{82.00} & \textbf{80.37} & \textbf{79.16} & 387.5 \\
    \bottomrule
    \end{tabular}
\end{table*}

\subsection{Loss Function}
The proposed row-wise detection network is supervised through two loss functions: lane existence loss $\mathcal{L}_{ext}$ and lane location probability loss $\mathcal{L}_{loc}$.
As we mentioned earlier, since the row-wise detection method is formulated as predicting the row-wise probability, both losses are constructed as sum of row-wise cross-entropy loss.
To be specific, the $\mathcal{L}_{ext}$ penalizes the network when it miss-classifies the existence of a certain lane class on the scene.
Let $\textbf{P}^{ext} \in \mathbb{R}^{N_{cls} \times H_{BEV} \times 2}$ be the lane existence ground truth where $N_{cls}$ is the number of lane classes, $H_{BEV}$ is the number of rows, and
$\textbf{p}^{ext}_{c,h} \in \mathbb{R}^2$ is a one-hot-encoded vector that indicates the existence of a lane of class $c$ on row $h$.
That is, $p^{ext}_{c,h,1} = 1$ if there exists a lane line of class $c$ on row $h$, and $p^{ext}_{c,h,0} = 1$ if the opposite is true.
Given the lane existence predictions $\hat{\textbf{P}}^{ext} \in \mathbb{R}^{N_{cls} \times H_{BEV} \times 2}$, we define the lane existence loss as,
\begin{equation}
    \mathcal{L}_{ext} = -\frac{1}{N_{cls} \cdot H_{BEV}} \sum_{c}^{N_{cls}}\sum_{h}^{H_{BEV}}\sum_{w}^{2}p^{ext}_{c, h, w} \cdot log (\hat{p}^{ext}_{c, h, i}).
\end{equation}

For the lane location loss, suppose that $\textbf{P}^{loc} \in \mathbb{R}^{N_{cls} \times H_{BEV} \times W_{BEV}}$ is the lane location probability ground truth, where $W_{BEV}$ is the number of columns.
$\textbf{p}^{loc}_{c,h}$ is a one-hot encoded vector that indicates the location of a lane of class $c$ on row $h$, that is, $p^{loc}_{c,h,w} = 1$ if there exists a lane line of class $c$ on row $h$ and column $w$, and $p^{loc}_{c,h,w} = 0$ if the opposite is true.
Given the lane location probability $\hat{\textbf{P}}^{loc} \in \mathbb{R}^{N_{cls} \times H_{BEV} \times W_{BEV}}$, we define the lane location probability loss as,

\begin{equation}
\small
    \mathcal{L}_{loc} = -\frac{1}{\sum_{c}^{N_{cls}}\sum_{h}^{H_{BEV}} p^{ext}_{c,h,1}} \sum_{c}^{N_{cls}}\sum_{h}^{H_{BEV}}\sum_{w}^{W_{BEV}} p^{loc}_{c, h, w} \cdot log(\hat{p}^{loc}_{c, h, w}) \cdot p^{ext}_{c, h, 1}.
\end{equation}

Unlike existence loss, the location loss is applied only to lane location probabilities for which its corresponding existence prediction is a positive flag.
This is because predicting the lane locations of non-existent lanes may have adverse effects to the training of the network.

Both existence and location loss are applied to both prediction results of the first and second stage heads.
Since both existence and location loss are normalized to each row (i.e., divided with $N_{cls} \cdot H_{BEV}$ and $\sum_{c}^{N_{cls}}\sum_{h}^{H_{BEV}} p^{ext}_{c,h,1}$, respectively), the total loss function is the summation of both lane existence loss and lane location probability loss,
\begin{equation}
    \mathcal{L}_{total} = (\mathcal{L}_{ext} + \mathcal{L}_{loc})_{1^{st} stage} + (\mathcal{L}_{ext} + \mathcal{L}_{loc})_{2^{nd} stage}.
\end{equation}

\section{Experiments}
In this section, we provide the implementation details that are required to reproduce our results.
Then we show and discuss our experimental results on the K-Lane dataset \cite{klane}.

\subsection{Dataset \& Metric}
We utilize the K-Lane dataset \cite{klane} throughout the experiments.
The dataset contains a collection of up to six labelled lane lines on over 15k frames of LiDAR point clouds, where the labels are provided in the form of BEV images.
Additionally, K-Lane includes carefully calibrated front camera images, which simplifies the visualization of LiDAR lane detection network output results as shown in Fig. \ref{fig3}.

For a fair comparison, we validate the performance with the F1-score, a harmonic mean of the precision and recall, following the previous work.
However, we measure the absolute computation of the network through floating point operations (FLOPs) rather than the inference speed as the speed measurements can vary depending on the quality of the hardware.
Thus, in the experiments, we assess precision, recall, and computational complexity, all of which are crucial for lane detection task.

\subsection{Implementation Details}
We implement our experiments using PyTorch \cite{pytorch} on an Ubuntu 18.04 machine with RTX3090 GPUs.
Unless stated otherwise, we set the maximum epochs to be 20 with a batch size of 4.
We use Adam \cite{adam} as the optimizer, with a learning rate of $10^{-4}$.
All of the experiments are conducted on the K-Lane dataset \cite{klane}.

\begin{table}[b]
    \centering
    \caption{Comparison of Performance for Various Hyper-parameters}
    \label{tab:exp_hyper}
    \begin{tabular}{ccccccc}
    \toprule
        \multicolumn{1}{c}{Backbone} & \multicolumn{3}{c}{2-Stage Head} & Overall & GFLOPs \\
        \cmidrule(lr){2-4}
        Depth & $T_{ext}$ & Depth & $W_{thick}$ & F1-score & \\
        \cmidrule(lr){1-6}
        1 & - & - & - & 79.52 & \textbf{379.64} \\
        3 & - & - & - & 82.37 & 385.08 \\
        5 & - & - & - & 82.25 & 390.52 \\
        \cmidrule(lr){1-6}
        2 & 0.3 & 1 & 5 & 82.57 & 384.76 \\
        3 & 0.3 & 1 & 5 & \textbf{82.74} & 387.48 \\
        5 & 0.3 & 1 & 5 & 82.57 & 392.92 \\
        \cmidrule(lr){1-6}
        3 & 0.3 & 3 & 5 & 82.39 & 387.70 \\
        3 & 0.3 & 5 & 5 & 82.63 & 387.90 \\
        \cmidrule(lr){1-6}
        3 & 0.5 & 1 & 5 & 82.49 & 387.40 \\
        3 & 0.7 & 1 & 5 & 82.54 & 387.24 \\
        \cmidrule(lr){1-6}
        3 & 0.3 & 1 & 3 & 82.71 & 387.44 \\
        3 & 0.3 & 1 & 7 & 82.52 & 387.54 \\
    \bottomrule

    \end{tabular}
\end{table}

\begin{figure}[t!]
\centering
\includegraphics[width=0.9\columnwidth]{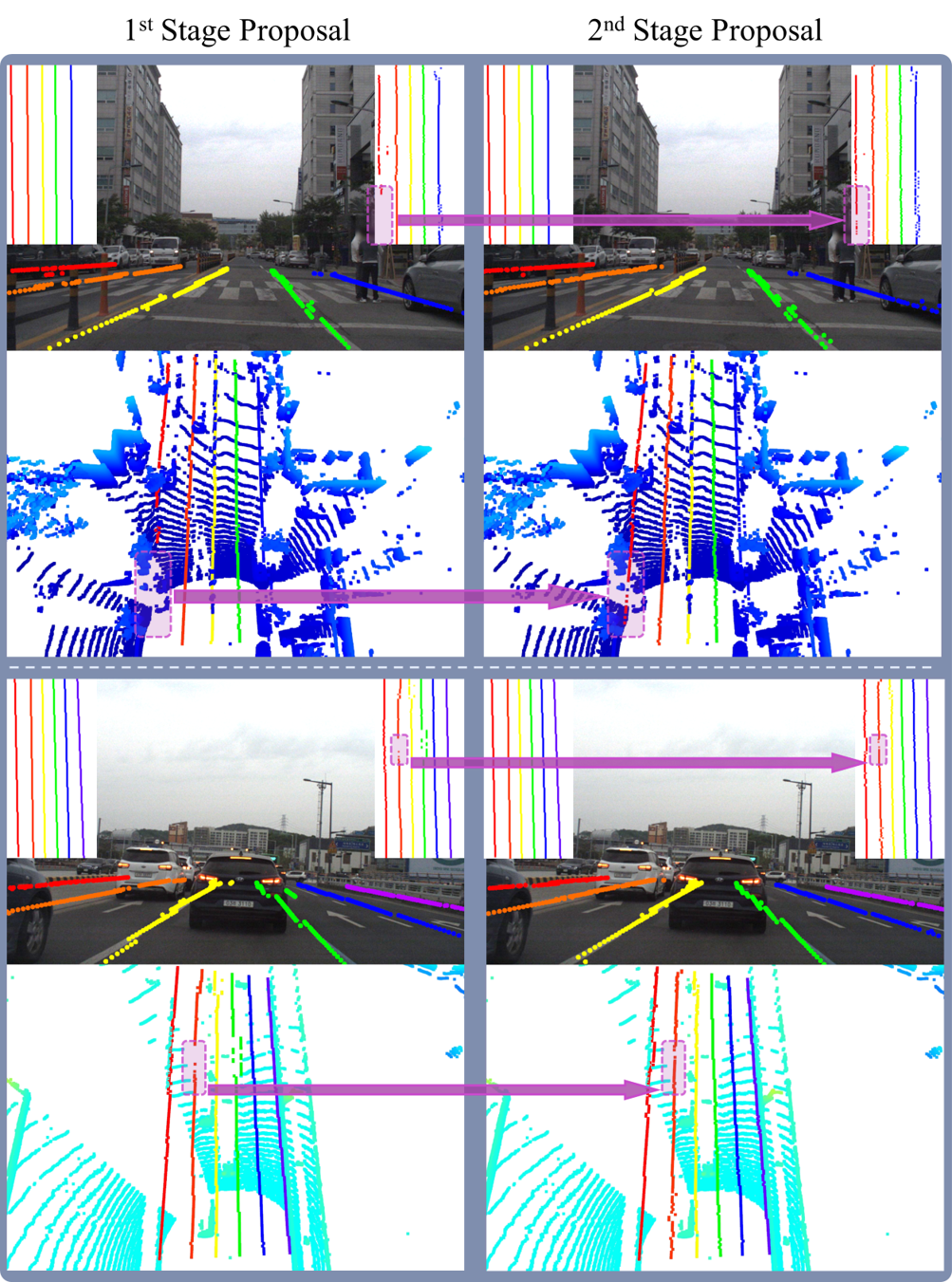}
\caption{Qualitative comparison of 1$^{st}$-stage and 2$^{nd}$-stage lane proposals from the proposed network. The 1$^{st}$-stage lane proposals is the intermediate output of the proposed network which is the input of 2$^{nd}$-stage detection head (i.e., Refinement head). We project the lane proposals onto the RGB camera image, where the BEV label is shown on the top-left corner, and the BEV prediction is shown on the top-right corner. We also visualize the lane proposals on the point cloud under its corresponding RGB camera image.}
\label{fig4}
\end{figure}

\subsection{Results on the K-Lane Dataset}
Fig. \ref{fig3} and Table \ref{tab:exp_perform} respectively show the qualitative and the quantitative results of our proposed network compared to the existing baseline, LiDAR lane detection network with global feature correlator (LLDN-GFC) based on segmentation approach \cite{klane}.
The row-wise detection network outperforms the existing segmentation-based network in terms of F1-score by +0.62 points.
Moreover, the row-wise detection network achieves state-of-the-art performance with less computational complexity, where the GFLOPs is about 30\% less when being compared to the existing baseline.
Though it is obvious that the overall performance of the model can be dependent on the hyper-parameter values as shown in Table \ref{tab:exp_hyper}, we note that the proposed network outperforms segmentation-based for most road conditions. In particular, when more than four lanes are occluded, the proposed network shows a significant improvement of 3.24\% over the existing SOTA neural network.
To the best of our knowledge, there is no publicly available LiDAR lane detection network other than the LLDN-GFC at the time of submission.

The superior performances of the proposed two-stage LiDAR lane detection network with row-wise approach may originate from {three} aspects.
First, the row-wise approach allows the detection head to directly produce the prediction values for all columns in the same row through the highly-optimized MLP operation.
This is in contrast with the segmentation-based approach, where the prediction values are obtained by applying $1 \times 1$ convolution to each column, which may result in a more inefficient operation.
Second, by applying the MLP per row, the detection head of the network may directly learn the correlation between all features on the same row in the feature map.
On the contrary, the each of the segmentation-based predictions is produced only from that specific grid in the feature map.
Third, the second-stage refinement through lane correlation enables the network to learn important connection between lane features globally, thus helping in the case of severe lane occlusions.
As shown in Fig. \ref{fig4}, the second-stage proposals, which reflects the correlation of the first-stage proposals as lane tokens, robustly detect the lanes (colorized in purple) which were not detected in the 1st-stage proposals due to vehicle occlusions.

Quantitatively, as shown in Table \ref{tab:exp_perform}, for the single-stage neural network, we observe only a 0.44\% improvement over prior SOTA network in the roads where more than 4 to 6 lanes are occluded. 
On the other hand, for the same occlusion level, the proposed two-stage network improves the SOTA by 3.24\%.

\subsection{Experiments on Design Choices}
There are several hyperparameter choices that affect the final performance of the proposed network.
On the first-stage network, the depth of the transformer-based global feature correlator (GFC-T) \cite{klane}, which partly reflect the model capacity, should be optimal w.r.t. the complexity of the problem and the number of samples in the dataset.
On the second-stage network, there are three hyperparameters: the lane existence threshold ($T_{ext}$), network depth, and the thickness of extracted lanes when collecting the lane tokens ($W_{thick}$).

To study the effect of each hyperparameter, we conduct several experiments with different hyperparameter values.
As seen in Table \ref{tab:exp_hyper}, all previously mentioned hyperparameters are affecting the network performance.

From the experimental results, we find that setting the backbone depth to be three yields the best performance in terms of F1-score.
Interestingly, the optimal depth for the second stage refinement head is found to be one, where deeper network lead to lower F1-score.
In addition, we observe that setting $W_{thick}$ to be five gives the optimal performance compared to other values.

We also perform ablation study on the effect of the second-stage network.
From Table \ref{tab:exp_perform}, we can see that the proposed two-stage row-wise detection network is especially robust on the occlusion cases, which empirically support our explanation on Section 3.

\section{Conclusions}
In this work, we have proposed a novel two-stage LiDAR lane detection network with row-wise detection approach.
The first stage network is responsible for predicting lane proposals, while the second stage network refines the first stage feature map through attention-based mechanism between lane tokens, before predicting the refined lane proposals.
From experimental results on the K-Lane dataset, the proposed network advances the state-of-the-art performance with an overall F1-score of 82.74\% by 0.62\% while reducing the GFLOPs by about 30\%.
In particular, the proposed network greatly improves the lane detection performance for severe occlusion cases.
Therefore, the proposed network enables a robust lane detection required for safe autonomous driving in the congested traffics where performance deterioration is previously significant.

\section*{ACKNOWLEDGMENT}

This work was partly supported by National Research Foundation of Korea (NRF) grant funded by the Korea government (MSIT)  (No. 2021R1A2C3008370).



\bibliographystyle{IEEEtran}
\bibliography{refer}

\end{document}